%% file: main.tex
\documentclass[a4paper,conference]{IEEEtran}

\usepackage[utf8]{inputenc} 
\usepackage[T1]{fontenc}    
\usepackage{hyperref}       
\usepackage{url}            
\usepackage{booktabs}       
\usepackage{amsfonts}       
\usepackage{nicefrac}       
\usepackage{microtype}      
\usepackage{xcolor}         
\usepackage{caption}
\usepackage{multirow}
\usepackage{graphicx}
\usepackage{enumitem}
\usepackage{amsmath}
\usepackage{soul}

\usepackage{balance}

\title{Deep Reinforcement Learning for Exact Combinatorial Optimization: Learning to Branch}

\author{%
  
  \IEEEauthorblockN{Tianyu Zhang\textsuperscript{\textsection}}
  \IEEEauthorblockA{University of Alberta\\
  Edmonton, Canada \\
  \texttt{tianyu.zhang@ualberta.ca}}
  \vspace{-0.30cm}
  \and
  \IEEEauthorblockN{Amin Banitalebi-Dehkordi}
  \IEEEauthorblockA{Huawei Technologies Canada Co., Ltd. \\
  Vancouver, Canada \\
  \texttt{amin.banitalebi@huawei.com}}
  \vspace{-0.30cm}
  \and
  \IEEEauthorblockN{Yong Zhang}
  \IEEEauthorblockA{Huawei Technologies Canada Co., Ltd. \\
  Vancouver, Canada \\
  \texttt{yong.zhang3@huawei.com}}
  \vspace{-0.30cm}
}

\begin{document}

\maketitle

\begingroup\renewcommand\thefootnote{\textsection}
\footnotetext{Work done during an internship at Huawei Technologies Canada Co., Ltd.}
\endgroup

\begin{abstract}
Branch-and-bound is a systematic enumerative method for combinatorial optimization, where the performance highly relies on the variable selection strategy. State-of-the-art handcrafted heuristic strategies suffer from relatively slow inference time for each selection, while the current machine learning methods require a significant amount of labeled data. We propose a new approach for solving the data labeling and inference latency issues in combinatorial optimization based on the use of the reinforcement learning (RL) paradigm. We use imitation learning to bootstrap an RL agent and then use Proximal Policy Optimization (PPO) to further explore global optimal actions. Then, a value network is used to run Monte-Carlo tree search (MCTS) to enhance the policy network. We evaluate the performance of our method on four different categories of combinatorial optimization problems and show that our approach performs strongly compared to the state-of-the-art machine learning and heuristics based methods.
\end{abstract}

\input{sections/introduction}

\input{sections/related-work}
\input{sections/background}
\input{sections/methodology}
\input{sections/result}
\input{sections/conclusion}

\bibliographystyle{unsrt}
\bibliography{common}

\end{document}

%% file: sections/introduction.tex
\section{Introduction}

\textit{Combinatorial optimization} is a broad topic covering several areas of computer science, operations research, and artificial intelligence. The fundamental goal of combinatorial optimization is to find optimal configurations from a finite discrete set that satisfy all given conditions, which involves enormous discrete search spaces. Examples include internet routing~\cite{climaco2007internet}, scheduling~\cite{crama1997combinatorial}, protein structure prediction~\cite{khimasia1997protein}, combinatorial auctions~\cite{sandholm2002algorithm}. Many real-life problems can also be formalized as combinatorial optimization problems, including the travelling salesman~\cite{larranaga1999genetic}, the vertex colouring~\cite{diaby2010linear}, and the vehicle routing problems~\cite{toth2002vehicle, golden2008vehicle}. As combinatorial optimization includes various NP-hard problems, there is a significant demand for efficient combinatorial optimization algorithms.

Several exact combinatorial optimization algorithms have been proposed to provide theoretical guarantees on finding optimal solutions or determining the non-existence of a solution. The core idea is to prune the candidate solution set by confidently introducing new conditions. Branch-and-bound (B\&B)~\cite{land1960automatic} is an example of an exact method to solve the combinatorial problem, which recursively divides the candidate solution set into disjoint subsets and rules out subsets that cannot have any candidate solutions satisfying all conditions. It has shown a reliable performance in the domain of mixed-integer linear programs (MILPs) to which many combinatorial problems can be reduced~\cite{lawler1966branch}. Several commercial optimization solvers (e.g. CPLEX, Gurobi) use a B\&B algorithm to solve MILP instances. However, two decisions must be made at each iteration of B\&B: \textit{node selection} and \textit{
variable selection}, which determine the next solution set to be partitioned, and the variable to be used as the partition rule, respectively. Most state-of-the-art optimizers use heuristics hard-coded by domain experts to improve the performance~\cite{gleixner2021miplib}. However, such heuristics are hard to develop and require adjustment for different problems~\cite{gasse2019exact}.

In recent years, an increasing number of studies have been focusing on training machine learning (ML) algorithms to solve MILP problems. The idea is that some procedural parts of the solvers may be replaced by ML models that are trained with historical data. However, most ML models are trained through supervised learning, which requires the mapping between training inputs and outputs. Since the optimal labels are typically not accessible, supervised learning is not capable for most MILP problems~\cite{bello2016neural}.
In contrast, reinforcement learning (RL) algorithms show a potential benefit to the B\&B decision-making, thanks to the fact that the B\&B decision-making process can be modelled as a Markov decision process (MDP)~\cite{gasse2019exact, mazyavkina2021reinforcement}. This offers an opportunity to use statistical learning for decision-making.

In this work, we provide an RL-based approach to learn a variable selection strategy, which is the core of the B\&B method. Our agent is trained to maximize the improvement of dual bound integral with respect to time in the B\&B method. We adopt the design of Proximal Policy Optimization (PPO)~\cite{schulman2017proximal}, combining the idea of imitation learning to improve the sample efficiency and advance imitated policy. We imitate the Full Strong Branching (FSB)~\cite{linderoth1999computational} variable selection rule to discourage the exploration of unpromising directions. We also introduce a Monte Carlo Tree Search (MCTS) like approach~\cite{sutton2018reinforcement} to encourage exploration during the training phase and reinforce the action selection strategy.

We evaluate our RL agent with four kinds of widely adopted combinatorial optimization problems. The experiments show that our approach can outperform state-of-the-art methods under multiple metrics. In summary, our contribution is threefold:
\begin{itemize}
    \item We implement and evaluate an RL-based agent training framework for B\&B variable selection problem and achieve comparable performance with the state-of-the-art GCNN approach using supervised learning.
    \item To facilitate the decision quality, we propose a new MDP formulation that is more suitable for the B\&B method.
    \item We use imitation learning to accelerate the training process of our PPO agent and propose an MCTS policy optimization method to refine the learned policy.
\end{itemize}


%% file: sections/related-work.tex
\section{Related Work\label{sec:relate}}
B\&B~\cite{land1960automatic} is one of the most general approaches for global optimization in nonconvex and combinatorial problems, which combines partial enumeration strategy with relaxation techniques. B\&B maintains a provable upper and lower (primal and dual) bound on the optimal objective value and, hence, provides more reliable results than heuristic approaches. However, the B\&B method can be slow depending on the selection of branching rules, which may grow the computational cost exponentially with the size of the problem~\cite{boyd2007branch}.

Several attempts have been made to derive good branching strategies. Current branching strategies can be categorized into hand-designed approaches that make selections based on heuristic scoring functions; and statistical approaches that use machine learning models to approximate the scoring functions. Most modern MILP solvers use hand-designed branching strategies, including  most infeasible branching~\cite{achterberg2005branching}, pseudocost branching (PC)~\cite{benichou1971experiments}, strong branching (SB)~\cite{lenstra2009traveling, linderoth1999computational}, reliability branching (RB)~\cite{achterberg2005branching}, and more. Strong branching provides the local optimal solution with the highest computational cost by experimenting with all possible outcomes. 
Pseudocost branching keeps a history of the success of performed branchings to evaluate the quality of candidate variables, which provides a less accurate but computationally cheaper solution. Reliability branching integrates both strong and pseudocost branching to balance the selection quality and time.

Given the fact that strong branching decisions provide a minimum number of explored nodes among all other hand-designed branching strategies but have a high computational cost, several studies have come up with the idea of approximating and speeding up strong branching strategies using statistical approaches. In \cite{alvarez2017machine}, a regressor is learned to predict estimated strong branching scores using offline data collected from similar instances. Similarly, a learning-to-rank algorithm that estimates the rank of variables can also provide reliable result~\cite{khalil2016learning, hansknecht2018cuts}, which is more reliable than mimicking the score function. However, these statistical approaches suffer from extensive feature engineering.

One common approach to reducing the feature engineering effort is to use the graph convolutional neural network (GCNN). Reference~\cite{dai2017learning} first proposed a GCNN model to solve combinatorial optimization problems, and reference~\cite{gasse2019exact} extended the structure to the context of B\&B variable selection, which is the closest line of work to ours. In~\cite{gasse2019exact}, authors show the GCNN can provide accurate estimation of strong branching with the shortest solving time in most of the considered instances. 

However, most recent statistical approaches for variable selection in B\&B use supervised learning techniques, which require a mapping between training inputs and expected labels. The quality of the model highly depends on the quality of training labels. As mentioned earlier, recent studies use strong branching scores or selections as training labels, which provides the local optimal solution, but is not guaranteed to be the global optimal solution. In general, we do not have access to optimal labels for most combinatorial optimization problems, and thus the supervised learning paradigm is not suitable in most cases~\cite{bello2016neural}. Another approach is to learn through the interactions with an uncertain environment and provide some reward feedbacks to a learning algorithm. This is also known as the reinforcement learning (RL) paradigm. The RL algorithm makes a sequence of decisions and learns the decision-making policy through trial and error to maximize the long-term reward. Previous studies have shown that the combinatorial optimization problem can be solved using RL algorithms, such as the travelling salesman problem~\cite{bello2016neural, dai2016discriminative, lu2019learning}, maximum cut problem~\cite{barrett2020exploratory, cappart2019improving, abe2019solving}, and more. This study proposes a deep reinforcement learning framework to learn the global optimal variable selection strategy. We adopt the structure of GCNN as the design of our policy and value network.

%% file: sections/background.tex
\section{Background}\label{sec:background}
In this section, we describe the fundamental concepts related to the paper, and provide formal definitions to various terms. 

\subsection{Mixed integer linear program (MILP)}
A MILP is a mathematical optimization problem that has a set of linear constraints, a linear objective function, and several decision variables that are continuous or integral with the form:
\begin{equation*}
    \operatorname*{\mathrm{arg\,min}}_{\mathbf{x}} \mathbf{c}^T\mathbf{x},\hspace{2em}
    \mathrm{s.t.}\hspace{2em}\mathbf{Ax}\leq\mathbf{b},\hspace{2em}\mathbf{l}\leq\mathbf{x}\leq\mathbf{u},
\end{equation*}
\vspace{-12pt}
\begin{equation*}
    x_i\in\mathbb{Z}\text{ where }i\in \mathcal{I},\hspace{2em}|\mathcal{I}|\leq n,
\end{equation*}
where $\mathbf{c}$ is the objective coefficient matrix, $\mathbf{x}$ is the variable vector, $\mathbf{A}\in\mathbb{R}^{m\times n}$ denotes the constraint coefficient matrix, $\mathbf{b}\in\mathbb{R}^{m}$ represents the constraint constant term vector, $\mathbf{l}\in\left(\mathbb{R}\cup\{-\infty\}\right)^n$ and $\mathbf{u}\in\left(\mathbb{R}\cup\{\infty\}\right)^n$ indicate the lower and upper variable bound vectors, respectively. Here $n$, $m$, and $\mathcal{I}$ respectively denote the number of variables, number of constraints, and index set of integer variables where $|\mathcal{I}|\leq n$. If a variable has no lower or upper bound, then we set the associated $l$ and $u$ to infinite values respectively. A \textit{candidate} solution is any assignment of $\mathbf{x}$ that satisfy the variable bounds. A \textit{feasible} solution is a candidate solution that satisfies all constraints in the MILP instance, and an \textit{optimal} solution is a feasible solution that minimize the objective function.

A MILP can be relaxed to a linear program (LP) by ignoring the integer constraints in the MILP; this is also called \textit{LP relaxation}. LP is convex and therefore can be solved efficiently using various algorithms, such as the simplex algorithm. Since removing the integer constraints expands the feasible set, the optimal solution for LP is then used as the lower bound for the corresponding MILP, namely the \textit{dual bound}.

\subsection{Branch-and-bound (B\&B) algorithm\label{sec:bnb}}
The B\&B algorithm constructs a search tree recursively. Each node in the search tree is a MILP. The B\&B algorithm can be described as follows. The original MILP is treated as the root node in the search tree. The algorithm then recursively picks a node from the search tree by a given node selection rule, picks a variable to decompose the selected node, and adds two children to the selected node that are produced by the decomposition. The dual bound of these two children are then being used to update the dual bound of the root node, and the algorithm selects the next node to expand. To decompose a MILP on variable $x_i$, we first find the optimal solution $\mathbf{x}^*$ to the LP relaxation. Then, if $\mathbf{x}^*_i$ does not meet the integrality constraint, we can decompose the MILP into two sub-problems with additional constraints $x_i\leq \left\lfloor x^*_i\right\rfloor$ and $x_i\geq \left\lceil x^*_i\right\rceil$. The variable $x_i$ is called the \textit{branching variable}, and all variables that can be selected are called \textit{branching candidates}.

\subsubsection{Strong branching (SB)}
SB is one of the most powerful state-of-the-art variable selection rules. The idea of SB is to test which branching candidate can provides the best improvement measured in children nodes. This method is a greedy method that selects the locally best variable to branch on, which usually works well in terms of the number of nodes visited to solve the problem. However, it requires to branch on every branching candidates to calculate the score, which is computationally expensive. Moreover, this greedy approach cannot guarantee to provide the global optimal selection.

\subsection{Markov decision process (MDP) formulation}
We can formulate the sequential decision making of variable selection as a MDP. Each node in the search tree can be encoded as a state. The agent exerts a branching variable from all branching candidates to decompose the current node. This action causes a transition to a child node. Through interactions with the MDP, the algorithm learns an optimal \emph{policy} $\pi$, that is a sequence of control actions starting from the root node.

\subsubsection{State}
\begin{figure}
    \centering
    \vspace{-8pt}
    \includegraphics[width=\columnwidth]{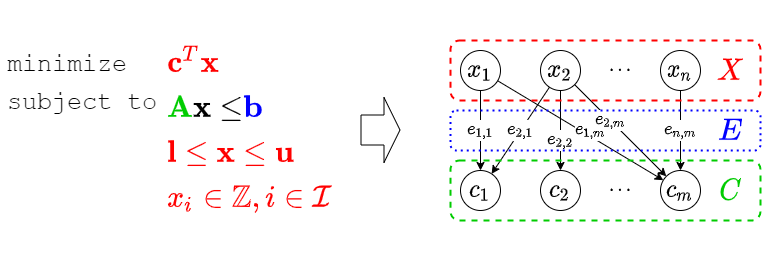}
    \vspace{-22pt}
    \caption{Bipartite graph representation $(X, E, C)$ of a MILP.}
    \label{fig:bipartite}
\end{figure}

The state $s_t$ at node $t$, can be represented as:
\begin{equation*}
    s_t = \{(X, E, C)_t, J_t\},
\end{equation*}
where the first tuple is the bipartite graph representation $(X, E, C)_t$ of the current node MILP, as done in \cite{gasse2019exact}, and index set $J_t$ is the index set of branching candidates. Two sets of nodes in the bipartite graph correspond to the $n$ variable to be optimized and $m$ constraints to meet. The edge $e_{i, j}$ is added if the variable $x_i$ has a non-zero coefficient $A_{i, j}$ in the constraint $c_j$, where $d_e$ features form the constraints constant term. $E\in\mathbb{R}^{m\times n\times d_e}$ represents the sparse edge feature matrix. $X\in\mathbb{R}^{n\times d_x}$ is a feature matrix for all variable nodes
, including the features extracted from the objective function and variable constraints.
Similarly, $C\in\mathbb{R}^{m\times d_c}$ represents the feature matrix for all constraint nodes, where each constraint 
is encoded into $d_c$ features. Figure~\ref{fig:bipartite} illustrates the bipartite representation of a general MILP instance. We calculate the optimal solution of the current node's LP relaxation and mark variables with integer constraints and having a non-integer solution as the branching variable to get $J_t$.

\subsubsection{Action and transition} 
The action at node $t$, denoted by $a_t$, determines the branching variable from the branching candidates: $a_t \in J_t$. After an action is performed, the search tree will add two children nodes to the current node and then prunes the search tree if needed, as described in Section~\ref{sec:bnb}. All children share the same $p(s_{t+1}|s_{t}, a_{t})$ in this study. 

\subsubsection{Reward}
The reward function is designed to encourage the agent increase the dual bound quickly with as less branching operations as possible. Because we do not control the selection of state, 
and the global dual bound is highly related to the search tree constructed based on the selection of branching node at each step, using the improvement of global dual bound is not valid as the selected branching node might not be able to improve the global dual bound by any action. Therefore, we calculate the reward based on the improvement of the local dual bound:
\begin{equation*}
    r(s_t, a_t) = \mathrm{min}\{\mathbf{c}^T\mathbf{x_{\lfloor t'\rfloor}}_{LP}^*, \mathbf{c}^T\mathbf{x_{\lceil t'\rceil}}_{LP}^*\}-\mathbf{c}^T\mathbf{x_t}_{LP}^*,
\end{equation*}
where $\mathbf{x_t}_{LP}^*$ is the dual bound of the current node $s_t$, $\mathbf{x_{\lfloor t'\rfloor}}_{LP}^*$ and $\mathbf{x_{\lceil t'\rceil}}_{LP}^*$ are respectively the dual bound of children nodes after adding constraint $x_{a_t}\leq \lfloor x_{a_t}^*\rfloor$ and $x_{a_t}\geq \lceil x_{a_t}^*\rceil$ to $s_t$.

%% file: sections/methodology.tex
\section{Methodology}\label{sec:methodology}
\begin{figure*}
    \centering
    \includegraphics[width=1.4\columnwidth]{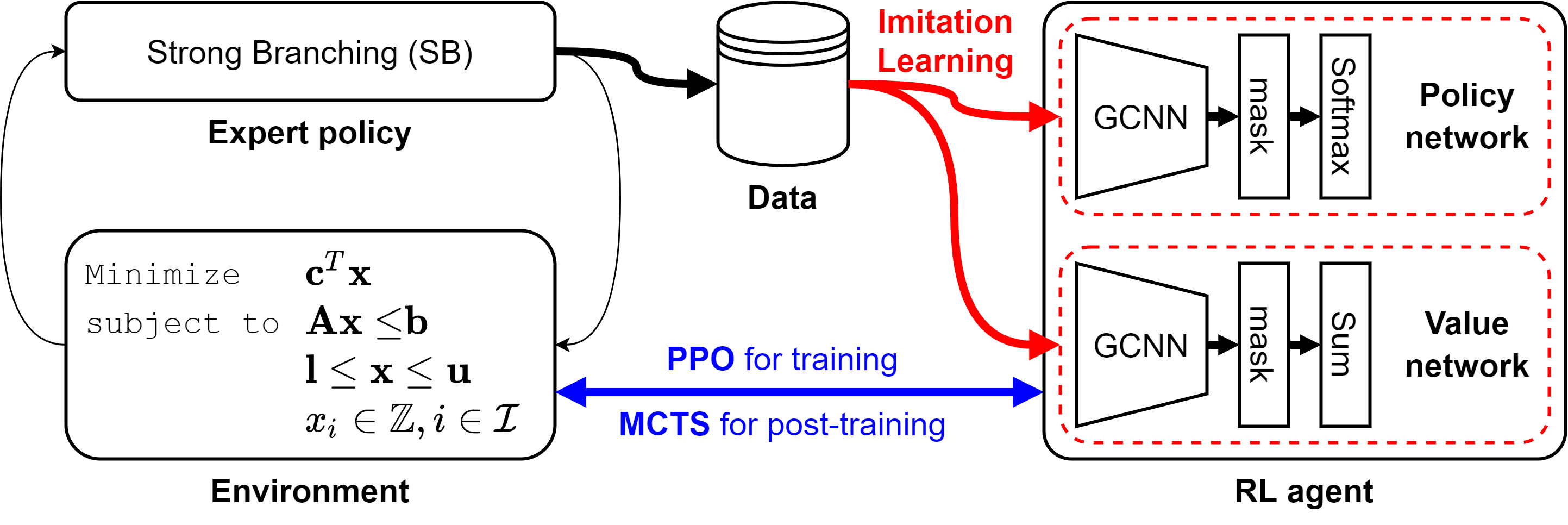}
    \vspace{-8pt}
    \caption{The flow-diagram of the proposed approach.}
    \label{fig:workflow}
\end{figure*}

In this section, we discuss the design of the RL agent, techniques to address the cold-start problem, the training algorithm, and how to exploit the knowledge of a trained RL agent to select branching variables from a given set of branching candidates. Figure~\ref{fig:workflow} shows an overview of our approach, which entails 
(1) designing the RL agent;
(2) using imitation learning to pre-train the RL agent;
(3) training the RL agent with PPO;
(4) finally, selecting reliable branching variables for test environments using RL agent based on the search result of Monte-Carlo tree search (MCTS); We describe each of these tasks below.

\subsection{Designing the RL agent}
Reinforcement learning methods can find a policy that maximizes the total reward, especially when the MDP is identified. In this study, we use a policy gradient-based method called proximal policy optimization (PPO) to find the optimal policy $\pi$ using the actor-critic framework. PPO has shown a strong performance in nearly all reinforcement learning tasks, thanks to the clipping method that limits the update of the behaviour policy within a trust region.

To evaluate the step size of the policy gradient method, PPO keeps tracking two policies, current policy $\pi_{\theta}$ and old policy $\pi_{\theta_{old}}$. Each policy contains two networks: a policy network that estimates the action distribution of a given state and a value network that estimates the state value. The state value $V(s_t)$ in this study is defined as follow:
\begin{equation*}
    V(s_t)=\sum_a p(a|s_t)\left(r(s_t, a) + \gamma \frac{V(\lfloor s_t'\rfloor) + V(\lceil s_t'\rceil)}{2}\right),
\end{equation*}
where $\gamma$ is a discount factor with value of 0.99 to encourage immediate reward, and states $\lfloor s_t'\rfloor$ and $\lceil s_t'\rceil$ are two children after branching on state $s_t$ with variable $x_a$. This is the different from the next state we obtained through the interaction with the environment. 
That is being said, state $s_t$ and $s_{t+1}$ might not have an edge in the search tree, because the node selection rule pick the next state from all leaf nodes in the search tree. In the calculation of the state value, the next state must be the child of the current state, to therefore correctly represent the state value in the search tree. If $s_t$ is a leaf node in the search tree, then the state value $V(s_t)$ is set to 0.

Because the state consists of a bipartite graph, we use graph convolutional neural network (GCNN) as our policy and value network. Previous studies also proved that GCNNs can effectively capture structural characteristics of combinatorial optimization problems. We adopt the similar GCNN design from \cite{gasse2019exact}, which use two successive passes to perform a single graph convolution. These passes are
\begin{equation*}
    c_p'\leftarrow f_\mathrm{c}\left(c_p, \sum_q^{(p, q)\in E} g_\mathrm{c}\left(\mathrm{emb}_\mathrm{x}(x_q), e_{p, q}, \mathrm{emb}_\mathrm{c}(c_p)\right)\right),
\end{equation*}
\begin{equation*}
    x_q'\leftarrow f_\mathrm{x}\left(x_q, \sum_p^{(p, q)\in E} g_\mathrm{c}\left(\mathrm{emb}_\mathrm{x}(x_q), e_{p, q}, c_p'\right)\right),
\end{equation*}
for all $p\in C, q\in X$. Next, the value of $x'$ is sent to a 2-layer perceptron. For the policy network, we apply masked softmax activation to estimate the action distribution. For the value network, we compute masked sum to predict the state value.

\subsection{Imitating the Strong Branching (SB)}
Theoretically, the RL agent can find the optimal policy $\pi$ from scratch after training for enough episodes. However, as the search tree is huge, with a branching factor usually over 1,000, training an RL agent from scratch becomes time-consuming and therefore impractical. To avoid the initial aimless exploration of the RL agent, we use the imitation learning approach to pretrain the RL agent policy and value network, paving the way for learning sophisticated policy. We select SB as our expert policy to generate offline training data, including the state, corresponding SB score for each branching candidate, as well as the reward. Then we reconstruct the state value $V(s_t)$ from the offline data and pretrain the policy and value network by minimizing the following loss:
\begin{equation*}
    L^{policy}(\theta)=-\frac{1}{N}\sum_{s, a\in\mathcal{D}}\mathrm{log}\pi_\theta(a|s)
\end{equation*}
\begin{equation*}
    L^{value}(\theta)=\frac{1}{N}\sum_{s\in\mathcal{D}}(V_\theta(s)-V(s))^2
\end{equation*}

\subsection{Training the RL agent}
Once the RL agent is pretrained using offline data, it is necessary to learn an advanced policy by interacting with the environment directly. To update the policy parameter $\theta$ with some trajectories generated through the interaction with the environment, we first save the parameters $\theta$ into $\theta_{{old}}$, and then calculate the loss as follows:
\begin{equation*}
    A_t=V(s_t)-V_{\theta}(s_t),\hspace{2em}r=\frac{\pi_\theta(a_t|s_t)}{\pi_{\theta_{old}}(a_t|s_t)},
\end{equation*}
\begin{equation*}
    L_t^{policy}(\theta)=\mathbb{E}_t\left[\mathrm{min}\left(rA_t, \mathrm{clip}(r, 1-\epsilon, 1+\epsilon)A_t\right)\right],
\end{equation*}
\begin{equation*}
    L_t(\theta)=\mathbb{E}_t\left[L_t^{policy}(\theta)-c_1A_t^2-c_2\sum_a \pi_\theta(a|s_t)\mathrm{log} \pi_\theta(a|s_t)\right],
\end{equation*}
where $V(s_t)$ is the state value calculated from the experiences, $V_{\theta}(s_t)$ is the value network estimated state value for $s_t$, and $\epsilon, c_1, c_2$ are hyperparameters of the model. In this study, we use $\epsilon=0.1, c_1=0.5, c_2=0.01$.

\subsection{Enhancing policy with Monte-Carlo Tree Search (MCTS)}
After we obtain the stable policy $\pi_\theta^*$ and $V_\theta^*$, it is essential to make reliable selections for a given state $s_t$. One common and straightforward approach is to take action with the highest estimated action probability, $\mathrm{argmax}_a\pi_\theta^*(a|s_t)$. However, this result could be biased when the policy is not optimized or has a significant variance. Since we have a tree-like search space, it is possible to adopt the idea of MCTS to update the policy $\pi_\theta^*$ further. In MCTS, we generate multiple trajectories starting from the current node. Then, we expand trajectories by taking action based on some probability distribution until a certain number of steps or the final state is reached. Finally, we pick the best action based on these trajectories. This is similar to the SB, except MCTS does not explore all branching candidates.

To run MCTS efficiently, we incorporate the knowledge learned by our RL agent. In the action selection step for MCTS, we use a modified version of upper confidence bound (UCB), which selects the action that maximizes the following equation:
\begin{equation*}
    \operatorname*{\mathrm{arg\,max}}_{a\in\mathcal{A}(s)}(s, a) + c\pi_\theta^*(a|s)\sqrt{\frac{\mathrm{log}(1+\sum_aN(s, a))}{N(s, a) + 1}}.
\end{equation*}
Here, the $Q(s, a)$ represents the action value based on the trajectories done previously, and $N(s, a)$ keeps tracking the number of times $a$ has been selected on state $s$. We introduce trained policy $\pi_\theta^*$ to encourage the algorithm to search for promising directions. To minimize the size of the search tree, we further limit the branching candidates on each state $s$ to $\mathcal{A}(s)$, which only contains the top $k$ actions based on the $\pi_\theta^*(s)$. In this study, we use $k=10$.

Also, to reduce the simulation time, we do not perform the branching operation when we run MCTS. Instead, we directly modify the constraint feature matrix and edge feature matrix to simulate the next state $s'$ based on the action, with half chance to reach the left child and half chance to reach the right child. We then set the reward of all actions to 0 and use the value network $V_\theta^*$ trained by the RL agent to calculate the value of $V_\theta^*(s')$, and use it to calculate the $Q(s, a)$. We initialize the action value and the visit count as follow:
\begin{equation*}
    Q(s, a)=\gamma V_\theta^*(s_t'),\hspace{2em}N(s, a)=1.
\end{equation*}
The $Q(s, a)$ and $N(s, a)$ are then updated when the agent reaches the leaf of the search tree or the maximum number of steps is reached. We apply the following update rule for each state $\{s_t, \cdots, s_0\}$:
\begin{equation*}
    Q(s_\tau, a_\tau)\leftarrow Q(s_\tau, a_\tau)+\frac{-Q(s_\tau, a_\tau)+\sum_{t'=\tau}^t\gamma^{t-\tau}V_\theta^*(s_t')}{N(s_\tau, a_\tau)+1}.
\end{equation*}
After all MCTS simulations are finished, we identify the action $a$ with the highest $Q(s, a)$ as the best branching variable for each state $s$ that has been visited at least ten times and use this to train the policy network by minimizing the cross-entropy loss. In this study, we limit the maximum depth to 3 and run 1,000 simulations of the MCTS for each state.

%% file: sections/result.tex
\begin{table*}[h]
    \centering
    \caption{Number of resulting B\&B nodes on the test data sets 
    \vspace{-8pt}
    \label{tab:node}}
        \begin{tabular}{ccccc}
        \toprule
        & \textbf{Set Covering} & \textbf{Independent Set} & \textbf{Combinatorial Auction} & \textbf{Capacitated Facility Location} \\ \midrule
        \textbf{Random} & 2225.20 & 257.60 & 25543.26 & 2292.24 \\
        \textbf{FSB} & 47.42 & 103.85 & \textbf{\textit{193.83}} &\textbf{\textit{47.9}}\\
        \textbf{GCNN} & 44.07 & \textbf{88.66} & 201.82 & 886.66\\
        \textbf{PPO-MCTS} & \textbf{\textit{43.90}} & 90.23 & 194.25 & 863.21\\ \bottomrule
        \end{tabular}
\end{table*}

\begin{table*}[h]
    \centering
    \caption{MILP instance solving time (in seconds) on the test data sets}\label{tab:time}
    \vspace{-8pt}
        \begin{tabular}{ccccc}
        \toprule
        & \textbf{Set Covering} & \textbf{Independent Set} & \textbf{Combinatorial Auction} & \textbf{Capacitated Facility Location} \\ \midrule
        \textbf{Random} & 19.2 & 15.36 & 99.08 & 92.28 \\
        \textbf{FSB} & 95.8 & 240.65 & 113.58 & 864.75\\
        \textbf{GCNN} & 3.28 & \textbf{\textit{6.54}} & 4.15 & 80.68\\
        \textbf{PPO-MCTS} & \textbf{\textit{3.13}} & 6.60 & \textbf{\textit{3.87}} & \textbf{\textit{77.24}}\\ \bottomrule
        \end{tabular}
\end{table*}

\begin{table*}[h]
    \centering
    \caption{Evaluation score on the test data sets\label{tab:reward}}
    \vspace{-8pt}
        \begin{tabular}{ccccc}
        \toprule
        & \textbf{Set Covering} & \textbf{Independent Set} & \textbf{Combinatorial Auction} & \textbf{Capacitated Facility Location} \\ \midrule
        \textbf{FSB} & 149930 & -191876 & -7093620 & 16119158\\
        \textbf{GCNN} & \textbf{150654} & \textbf{-191123} & -7077028 & 16159789\\
        \textbf{PPO-MCTS} & 150652 & -191139 & \textbf{-7076023} & \textbf{16160324}\\ \bottomrule
        \end{tabular}
\end{table*}

\begin{table*}[h]
    \centering
    \caption{Performance comparison between PPO and PPO-MCTS\label{tab:ablation}}
    \vspace{-8pt}
        \begin{tabular}{ccccccc}
        \toprule
        & \multicolumn{2}{c}{\textbf{Nodes visit}} & \multicolumn{2}{c}{\textbf{Solving time}} & \multicolumn{2}{c}{\textbf{Evaluation score}} \\\cmidrule(lr){2-3}\cmidrule(lr){4-5}\cmidrule(lr){6-7}
        & PPO & PPO-MCTS & PPO & PPO-MCTS & PPO & PPO-MCTS\\\midrule
        \textbf{Set Covering}                   & 57.68 & \textbf{43.90} & 3.52 & \textbf{3.13} & 150651 & \textbf{150652} \\
        \textbf{Independent Set}                & \textbf{88.18} & 90.23 & 8.34 & \textbf{6.60} & -203328 & \textbf{-191139} \\
        \textbf{Combinatorial Auction}          & 270.97 & \textbf{194.25} & 5.28 & \textbf{3.87} & -7077229 & \textbf{-7076023} \\
        \textbf{Capacitated Facility Location}  & 2202.46 & \textbf{863.21} & 139.65 & \textbf{77.24} & 16155103 & \textbf{16160324} \\\bottomrule
        \end{tabular}
\end{table*}

\section{Evaluation\label{sec:result}}
In this section, we study the efficacy of different variable selection strategies. We adopt the average solving time, average number of resulting B\&B nodes, and average dual integral as our evaluation metrics. All experiments are repeated five times with different random seeds to eliminate randomness. All numbers are the averaged value across all five runs.

\subsection{Data sets}
To test the generalizability of our framework, we evaluate our approach on four different types of NP-hard problems. The first problem is called set covering problem proposed in~\cite{balas1980set}. Our instances contain 1,000 columns and 500 rows per instance. The second problem is generated following the arbitrary relationships procedure of~\cite{leyton2000towards}. This problem is also known as the combinatorial auction problem. In our experiment, we generate instances with 100 items for 500 bids. Our third data set is called capacitated facility location described in~\cite{cornuejols1991comparison}. We collect instances with 100 facilities and 100 customers. The last data set we used in this study is proposed in~\cite{bergman2016decision}, which is called the maximum independent set problem. The affinity is set to 4, and the graph size is set to 500 in this study.

These problems are selected based on the previous works and the hardness of the problem itself. According to~\cite{gasse2019exact}, these problems are the most representative of the types of integer programming problems encountered in practice. We use SCIP 7.0.3~\cite{gamrath2020scip} as the backend solver throughout the study, with ecole 0.7.3~\cite{prouvost2020ecole} as the environment interface. All SCIP parameters are kept to default in this study.

\subsection{Baselines}
In the rest of this paper, we use PPO-MCTS to refer to our proposed reinforcement learning framework. We compare our approach with three different variable selection baseline strategies. The first naive baseline strategy is the pure random strategy, in which we select the branching variable from a set of candidate variables uniformly. We use the full strong branching (FSB) strategy as our second baseline, which we use the default parameter defined in SCIP in this study. We also re-implemented the GCNN model from~\cite{gasse2019exact} as our third baseline. Based on the ML4CO NeurIPS 2021 competition result, the GCNN model yields the best performance among all other competing methods~\cite{ml4co,ml4co_Nuri,ml4co_banitalebi}. The performance of PPO agents that have no MCTS learning afterwards (PPO) is also reported for ablation study.

\subsection{Evaluation metrics}
We evaluate the performance of each approach using three metrics, including the average solving time for each problem instance, the average number of B\&B search tree nodes visited before the problem is solved, and a reward score that takes into account both the solving time and the improvement of the dual bound. Solving time and the number of nodes visited measure the computational cost of each algorithm. Solving time is evaluated based on the wall clock time, including feature extraction time, model inference time, branching time, and more. Therefore, a shorter solving time does not guarantee to optimize the number of nodes visited during the B\&B method. To optimize the branching variable selection strategy, we expect to minimize the number of nodes visited during the branching and the total solving time to select optimal branching variables with minimum computational cost. The score is calculated by:

\begin{equation}
    -T\mathbf{c}^T\mathbf{x}^*+\int_{t=0}^T\mathbf{z}^*_t\partial t,
\end{equation}
where $\mathbf{x}^*$ is the optimal solution of the MILP instance, $T$ is the time budget to solve the problem, and $\mathbf{z}^*_t$ is the best dual bound at time $t$. This score is to be maximized, representing a fast improvement of the dual bound. This reward metric was first introduced in the ML4CO NeurIPS 2021 competition~\cite{ml4co} and is expected to be adopted further by the community.

\subsection{Experiment result}
Table~\ref{tab:node} shows the average number of nodes visited before solving the instances for each approach. We noticed both GCNN and our PPO-MCTS have more nodes visited in complex problems, namely the combinatorial auction and capacitated facility location problems, compared to FSB. We conclude this to the fact that the number of branching candidates in these two problems are more significant than the other two problem types, and therefore leads to the approximate function getting more complex, which lowers the performance of the GCNN model. Similarly, as the search tree branching factor grows, RL agents become more challenging to learn the environment thoroughly. The agent can potentially struggle in local optimal as the maximum depth is set to three for our PPO-MCTS agent. It is worth noting that number of nodes on its own is not enough of a measure to judge different approaches with. There reason is that a method may visit a larger number of nodes, but may in fact be faster on each visit and result a better overall reward value for convergence.

On the other hand, it is readily seen from Table~\ref{tab:time} that FSB takes a significant amount of time to select a variable for each node in these two problems, and therefore the total solving time for FSB is the longest compared to all other approaches in all four problems. The GCNN and PPO-MCTS are having similar inference time as the network designs are similar. As the PPO-MCTS has a lower average number of visited nodes in all but independent set problems, our method provides the shortest solving time in all problems except the independent set problem. However, the performance differences between GCNN and PPO-MCTS on independent set problem in all three metrics are negligible, which proves the effectiveness of our proposed framework. In addition, when the problem is easy, such as the set covering and independent set problems, both GCNN and PPO-MCTS can find better branching variables with fewer nodes visited than FSB. In general, PPO-MCTS has a slightly better performance across different data sets and metrics than GCNN, with a trade-off on the training expenses.

The average evaluation score for each approach is shown in Table~\ref{tab:reward}. 
The GCNN and PPO-MCTS approaches keep dominating the score in all problems, whereas the PPO-MCTS has a higher score on challenging problems, and GCNN performs better with easy problems. 

\subsection{Ablation study}
We present an ablation study of the proposed PPO-MCTS model to evaluate the importance of having post MCTS retraining. Table~\ref{tab:ablation} demonstrates the performance of PPO without post MCTS retraining on all metrics across all data sets, denoted by PPO, comparing with the proposed PPO-MCTS. It is observed that the PPO-MCTS perform better than PPO in all cases, except the number of nodes visited in the independent set problem. This empirical result suggests that the post MCTS retraining offers a better performing RL agent.

%% file: sections/conclusion.tex
\section{Conclusion\label{sec:conclusion}}
We proposed a reinforcement learning framework to learn the variable selection policy for the B\&B method. We formulated a reward function that helps the agent learn optimal policies without generating labels. We used imitation learning and MCTS to deal with sample inadequacy challenges by initializing the policy to a relatively good policy and enhancing it with multiple steps look-ahead. We demonstrated the performance of the proposed framework with three baseline approaches on four NP-hard problems and showed that the proposed method yields a strong performance in most problems.


\balance